\documentclass{article}

\usepackage[final]{corl_2026}
\usepackage[utf8]{inputenc}
\usepackage[T1]{fontenc}
\usepackage{amsmath,amssymb,amsthm,mathtools}
\usepackage{booktabs,tabularx,array,longtable}
\usepackage{graphicx,float}
\usepackage{microtype}
\usepackage{enumitem}
\usepackage{xcolor}
\usepackage{placeins}

\newtheorem{theorem}{Theorem}
\newtheorem{proposition}{Proposition}
\newcommand{\R}{\mathbb{R}}
\newcommand{\eps}{\varepsilon}

\DeclareMathOperator{\spec}{spec}
\DeclareMathOperator{\RePart}{Re}

\title{One Trajectory, Two Requirements: Accuracy and  Local Stability in Euler Sampling for Flow Matching Policies}

\author{
Mahindra Rautela\\
Los Alamos National Laboratory\\
\texttt{mrautela@lanl.gov}
}

\begin{document}
\maketitle

\begin{abstract}
Flow matching robot policies commonly use a fixed Euler step count selected from offline accuracy sweeps. At inference time, however, each policy input (visual observation, robot state, and language instruction) together with a source-noise draw defines a different ODE trajectory and sequence of local Jacobians. We ask whether a step count that accurately follows the nominal noise-to-action trajectory can still be too coarse for strongly contracting directions around it. An exact two-coordinate system and a controlled augmentation of six trained flow models show that trajectory accuracy and local Euler stability can impose different step requirements even when the original learned transport is unchanged. We then study a frozen $\pi_{0.5}$-LIBERO checkpoint without modifying its weights or action dynamics. Across 30 policy inputs from 10 LIBERO-10 tasks, the Jacobian exhibits strong contracting directions near the action endpoint, yet no sampled mode violates the Euler non-expansion condition. On five of those inputs, accuracy sweeps require 4--9 Euler steps at 5\% terminal error and 87--112 at 1\%. These five paired cases are therefore consistent with accuracy-limited rather than stability-limited sampling under this local criterion, while the controlled experiments show why endpoint accuracy alone cannot rule out a separate stability requirement.
\end{abstract}


\section{Introduction}
\label{sec:intro}

\emph{Flow matching policies for robot control.}
Flow matching is increasingly used for continuous action generation in robot policies. Recent SOTA vision--language--action (VLA) models such as $\pi_0$ and $\pi_{0.5}$ use a flow matching action expert to generate action chunks conditioned on visual observations, robot state, and language instructions~\cite{black2024pi0,black2025pi05}. Related flow-based policies have also explored reducing the number of sampling steps for faster action generation~\cite{zhang2025flowpolicy,dai2025actionflow}. Because these policies generate actions by numerically integrating a learned flow, the choice of integration step count becomes part of the inference procedure.

\emph{Fixed Euler step budgets.}
For flow matching policies, actions are generated by integrating a learned ordinary differential equation (ODE) from source noise to an action. The $\pi_0$ family, for example, uses a flow matching action expert that maps noise to an action chunk~\cite{black2024pi0,black2025pi05}. Each Euler step requires another evaluation of the action expert, so fewer steps directly reduce inference cost. In practice, the step count is often selected by sweeping several values on a fixed evaluation set and choosing one that gives acceptable endpoint accuracy.

At inference time, each policy input (visual observation, robot state, and language instruction) together with the source-noise draw defines a different noise-to-action trajectory. Along this trajectory, the state Jacobian $J_v=\partial v_\theta/\partial x$ describes how nearby states locally contract or expand. Because these local dynamics vary across trajectories, a step count chosen only from offline endpoint-accuracy sweeps may not be sufficient for every inference trajectory.

\emph{Accuracy and stability answer different questions.}
Trajectory accuracy asks whether the nominal Euler path ends close to a refined numerical solution of the learned ODE. Local stability asks what Euler does to nearby perturbations. At a point on the path, the state Jacobian can contain a direction that the continuous flow contracts rapidly. If the Euler step is too large, that same direction can be amplified instead. A trajectory can therefore be accurate at its endpoint while the local Euler update misrepresents how nearby states should behave. Conversely, a large Jacobian does not automatically force a larger step count: the contraction rate must be compared with the actual timestep used by the sampler.

\paragraph{Research question, scope, and contributions.}
We ask: \emph{for a frozen flow matching policy and a stated numerical tolerance, is the Euler step budget set by trajectory accuracy alone, or can local stability impose an additional requirement?}
We study this question in three stages. First, an exact two-coordinate system shows that accuracy and stability can require different Euler budgets. Second, a controlled contraction added to trained flow matching models changes the stability requirement without changing the original learned transport. Third, we derive a grid-aware Euler stability condition and evaluate it on a frozen $\pi_{0.5}$-LIBERO checkpoint. Across 30 policy inputs from 10 LIBERO tasks, we find strong local contraction but no sampled stability violation. On five inputs with paired accuracy sweeps, the step budget is instead set by trajectory accuracy under our criterion.

\begin{figure}[t]
\centering
\includegraphics[width=\linewidth]{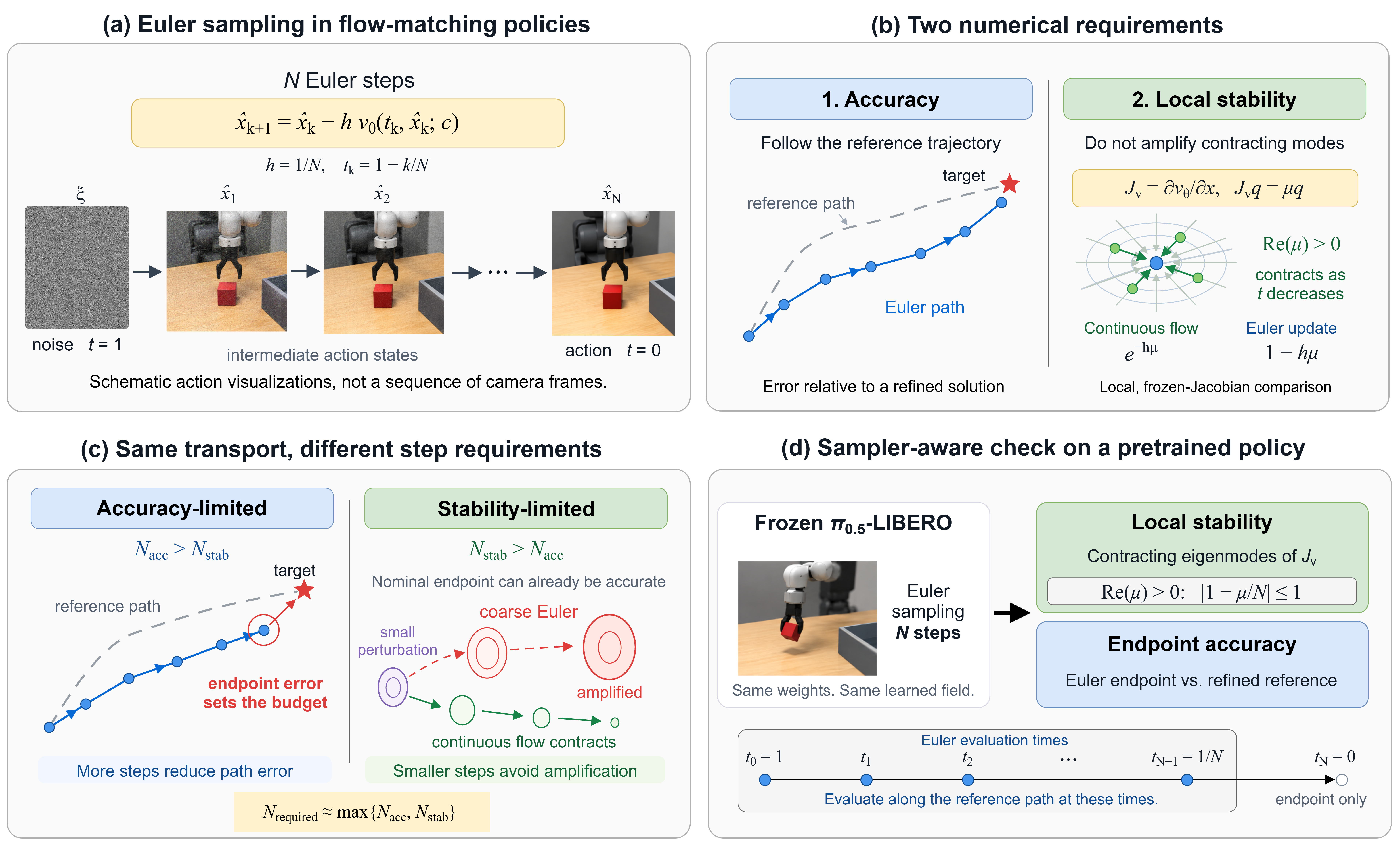}
\vspace{-4mm}
\caption{Accuracy and local stability impose distinct numerical requirements in Euler sampling. (a) Flow matching policy sampling from noise to action with \(N\) Euler steps. (b) Accuracy measures deviation from a refined reference trajectory, while local stability measures amplification of contracting Jacobian eigenmodes. (c) Either requirement can determine the necessary step budget. (d) For a frozen pretrained policy, we compare endpoint accuracy with a grid-aware local stability measure.}
\vspace{-2mm}
\label{fig:stiff_flow}
\end{figure}

\section{Related Work}
\label{sec:related}

\textbf{Generative policies and few-step action generation.}
Diffusion Policy generates conditional action sequences through iterative denoising \citep{chi2023diffusion}, while the $\pi_0$ family uses flow matching action experts inside pretrained VLA systems \cite{black2024pi0,black2025pi05}. FlowPolicy trains consistency across flow states for one-step 3D-conditioned action generation \citep{zhang2025flowpolicy}, and Coda adds a learned endpoint correction to a frozen few-step policy \citep{tang2026coda}. AdaFlow adapts Euler step sizes using conditional velocity-target variance \citep{hu2024adaflow}, while ProbeFlow uses velocity-direction information to select an inference budget and compares several integration schemes \citep{fang2026probeflow}. These works motivate efficient action generation. We keep the learned field and uniform Euler solver fixed and ask which numerical requirement makes a given step budget admissible.

\textbf{Numerical analysis of learned flows.}
Flow matching and neural ODEs place numerical integration directly inside inference \citep{lipman2023flow,chen2018neural}. SharpEuler allocates Euler evaluation times from an offline acceleration profile \citep{gupta2026sharpen}. STORK uses stabilized Runge--Kutta ideas for stiff diffusion and flow ODEs \citep{tan2026stork}. Tao and Choi analyze Jacobian strain, vorticity, and error propagation, while Xiao compares classical solvers and learned Jacobian eigenvalues \citep{tao2026strain,xiao2026euler}. We use classical absolute-stability ideas \citep{hairer1996solving} to separate trajectory accuracy from non-expansion of locally contracting eigenmodes, and explicitly avoid treating a large Jacobian as sufficient evidence of a stability bottleneck.

\section{Accuracy and Local Stability in Euler Sampling}
\label{sec:framework}

We first work directly in the native OpenPI sampling time. We denote the fixed policy conditioning by $c$: the visual observation, robot state (proprioception), and language instruction supplied at one replanning query. The Gaussian source-noise tensor $\xi$ is kept separate because it provides the initial action-flow state. For fixed $c$, the learned field is
\begin{equation}
\frac{dx}{dt}=v_\theta(t,x;c), \qquad t:1\rightarrow0,
\label{eq:native_ode}
\end{equation}
where $t=1$ is source noise and $t=0$ is the final action. This is the convention used by the $\pi$ models whereas standard flow matching typically uses the opposite convention, with $t=0$ for the source distribution and $t=1$ for the data distribution. With $h=1/N$, uniform Euler uses
\begin{equation}
 t_k=1-k/N,\qquad
 \widehat x_{k+1}=\widehat x_k-h\,v_\theta(t_k,\widehat x_k;c).
\label{eq:native_euler}
\end{equation}

Sampling time is not the robot's physical execution time. The state $x\in\R^d$ denotes the entire internal action-flow state that is iteratively
transported from noise to the final action representation. For the $\pi_{0.5}$-LIBERO checkpoint studied later, this state is a $10\times32$ tensor, so $d=320$ (action-chunk length of 10 $\times$ internal coordinates per action of 32). Thus, each Euler step updates the full action-flow state jointly, rather than advancing the robot by one physical control step. Fig.~\ref{fig:stiff_flow} summarizes the overall framework, highlighting the distinct accuracy and local-stability requirements in Euler sampling.

\subsection{Accuracy follows the nominal path}
A refined numerical solution defines the reference endpoint $x^{\rm ref}_{end}(c,\xi)$ for fixed conditioning input $c$ and source noise $\xi$. For a tested Euler budget $N$, we measure
\begin{equation}
 E_N(c,\xi)=
 \frac{\|\widehat x_{end}^{(N)}(c,\xi)-x^{\rm ref}_{end}(c,\xi)\|_2}
 {\|x^{\rm ref}_{end}(c,\xi)\|_2}.
\label{eq:per_context_error}
\end{equation}
Since endpoint error may not decrease consistently with the number of steps, we require the tolerance to hold for all larger tested step counts. For a finite tested budget set $\mathcal B$, the persistent accuracy threshold is
\begin{equation}
 N_{\rm acc}^{\mathcal B}(c,\xi;\eps)=
 \min\{N\in\mathcal B:E_n(c,\xi)\le\eps\ \text{for every tested }n\ge N\}.
\label{eq:per_context_nacc}
\end{equation}
This is a \emph{trajectory-level} quantity. An aggregate validation error can be useful for choosing a global budget, but it can hide variation across individual policy inputs and noise draws.

Along the exact path, the velocity changes with the material acceleration 
\begin{equation}
 a_\theta(t,x)=\partial_t v_\theta(t,x)+J_v(t,x)v_\theta(t,x),
 \qquad J_v=\frac{\partial v_\theta}{\partial x}.
\label{eq:material_acc}
\end{equation}
The local backward-time expansion is
\begin{equation}
 x(t-h)=x(t)-h v_\theta(t,x(t))+\tfrac12 h^2 a_\theta(t,x(t))+O(h^3),
\end{equation}
so Euler omits the leading $O(h^2)$ term. We measure terminal error directly rather than infer $N_{\rm acc}$ from acceleration alone \cite{gupta2026sharpen}.

\subsection{Stability follows nearby perturbations}
Accuracy follows one nominal state. Stability asks what happens to a nearby state \cite{hairer1996solving}. Let $\delta\in\R^d$ be an infinitesimal perturbation at a point $x(t)$ on the trajectory. Linearization gives
\begin{equation}
 \frac{d\delta}{dt}=J_v(t,x(t);c)\,\delta,
 \qquad J_v\in\R^{d\times d}.
\label{eq:perturb}
\end{equation}
If $J_v q=\mu q$, then a perturbation along the eigenvector $q$ evolves over one backward-time interval $h$ as $e^{-h\mu}q$. Because sampling moves from larger to smaller $t$, a mode with $\RePart(\mu)>0$ contracts continuously. Euler instead multiplies that mode by $1-h\mu$. Requiring Euler not to amplify a continuously contracting mode gives
\begin{equation}
 |1-h\mu|\le1
 \quad\Longleftrightarrow\quad
 N\ge \frac{|\mu|^2}{2\RePart(\mu)}.
\label{eq:native_mode}
\end{equation}
Define the local reverse-time stability scale
\begin{equation}
 \chi_{\leftarrow}(t,x;c)=
 \max_{\mu\in\spec(J_v),\ \RePart(\mu)>0}
 \frac{|\mu|^2}{2\RePart(\mu)},
\label{eq:reverse_chi}
\end{equation}
with $\chi_{\leftarrow}=0$ when no such mode exists. The Jacobian is evaluated \emph{on} the nominal path but describes directions \emph{around} it. An eigenvector can mix many state coordinates; it is not generally one action dimension.

This is a local stability test at a fixed point along the trajectory. It is exact for the scalar and diagonal systems below, but in general it does not guarantee that finite perturbations will contract in coupled, time-varying dynamics \citep{trefethen2005spectra}.

\subsection{The joint step requirement}
The conceptual comparison for one fixed conditioning input and source-noise draw is
\begin{equation}
 N_{\rm required}(c,\xi;\eps)
 \approx \max\{N_{\rm acc}(c,\xi;\eps),N_{\rm stab}(c,\xi)\}.
\label{eq:joint_philosophy}
\end{equation}
The approximation sign is deliberate: outside the controlled systems below, the stability measurement uses finitely sampled local eigenmodes and therefore does not characterize all finite perturbations. We call a case \emph{stability-limited} only when a checked Euler stability requirement excludes a budget already accepted by the stated accuracy criterion.

\section{Controlled Experiments Separating Accuracy and Stability}
\label{sec:controlled}

We first ask whether accuracy and local stability can require different numbers of Euler steps. We study this question using an exactly solvable system, followed by trained flow matching models.

\subsection{Experiment 1: an exact two-coordinate system}
For this controlled proof only, let $s=1-t$ denote forward sampling time. Consider
\begin{equation}
 \dot x_1=1+\beta s,\qquad \dot x_2=-\kappa x_2,
 \qquad \beta\ge0,\ \kappa>0.
\label{eq:toy}
\end{equation}
The first coordinate controls trajectory accuracy: increasing $\beta$ makes the velocity vary more along the nominal path. The second coordinate controls local stability: increasing $\kappa$ strengthens a contracting direction around that path without changing the first-coordinate motion.

\begin{theorem}[Exact joint budget]
\label{thm:exact}
For $\eps>0$ and uniform Euler on $s\in[0,1]$, terminal accuracy $|x_1(1)-\widehat x_{1,N}|\le\eps$ and stepwise non-expansion of $x_2$ hold simultaneously if and only if
\begin{equation}
 N\ge N^*(\eps)=\max\left\{\left\lceil\frac{\beta}{2\eps}\right\rceil,
                                  \left\lceil\frac{\kappa}{2}\right\rceil\right\}.
\label{eq:exact_budget}
\end{equation}
\end{theorem}
Euler's first-coordinate terminal error is exactly $\beta/(2N)$, whereas a perturbation in the second coordinate is multiplied by $1-\kappa/N$ at each step. Thus $\beta$ and $\kappa$ independently vary the accuracy and stability requirements.

We test the theorem with 1000 initial conditions, $x_1(0)\sim\mathcal N(0,1)$ and $x_2(0)\sim\mathcal N(0,10^{-12})$, using $\eps=0.2$ and every integer $N=1,\ldots,128$. The four parameter pairs $(\beta,\kappa)\in\{0.1,10\}\times\{1,50\}$ produce easy, accuracy-limited, stability-limited, and jointly limited regimes. In addition to terminal error, we track the maximum transverse amplification $R_z(N)$; its exact expression is given in Appendix~\ref{app:controlled}, and $R_z(N)>1$ exactly when $N<\kappa/2$.

\begin{table}[t]
\centering\small
\caption{Exact mechanism separation at $\eps=0.2$. Accuracy and stability independently set the admissible Euler budget, and the observed sweep matches the analytic thresholds in every regime.}
\label{tab:analytic}
\begin{tabular}{@{}lrrrr@{}}\toprule
Regime & $(\beta,\kappa)$ & $N_{\rm acc}^*$ & $N_{\rm stab}^*$ & $N^*$\\\midrule
Easy & $(0.1,1)$ & 1 & 1 & 1\\
Accuracy-limited & $(10,1)$ & 25 & 1 & 25\\
Stability-limited & $(0.1,50)$ & 1 & 25 & 25\\
Jointly limited & $(10,50)$ & 25 & 25 & 25\\\bottomrule
\end{tabular}
\end{table}

\begin{figure}[t]
\centering
\includegraphics[width=\linewidth]{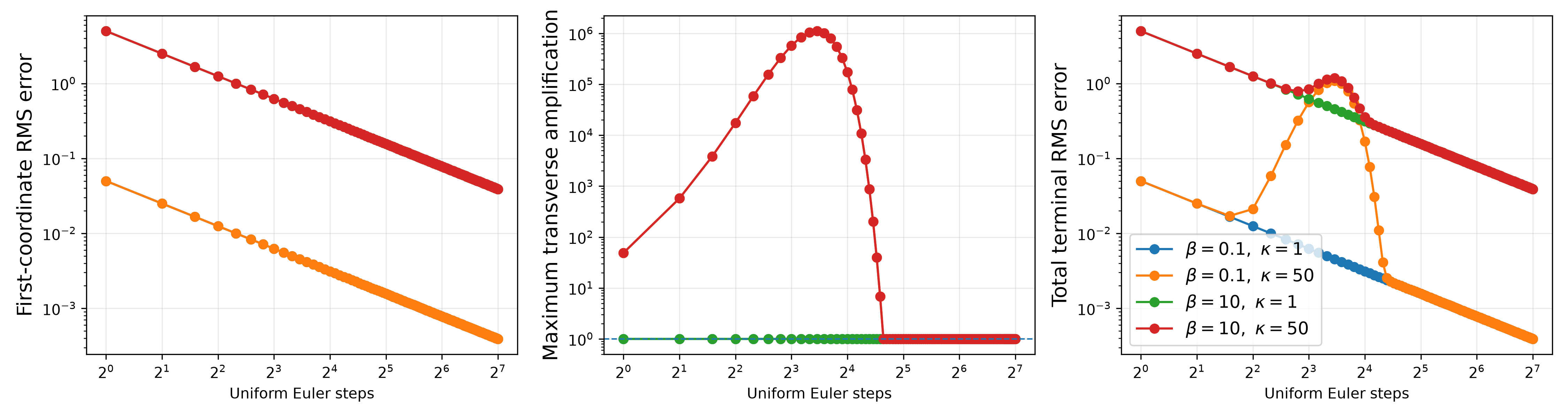}
\vspace{-4mm}
\caption{\textbf{Experiment 1: exact mechanism separation.} Left: first-coordinate terminal error decreases with the Euler budget and is controlled by $\beta$. Middle: transverse amplification is controlled by $\kappa$ and becomes non-expansive exactly at $N=\lceil\kappa/2\rceil$. Right: the total terminal error combines both effects. Increasing $\beta$ therefore changes the accuracy requirement, while increasing $\kappa$ moves a separate, sharp Euler stability boundary.}
\vspace{-3mm}
\label{fig:exact_mechanism}
\end{figure}

Table~\ref{tab:analytic} and Fig.~\ref{fig:exact_mechanism} confirm the theoretical step requirements in all four cases. The clearest example is the case with low acceleration and strong contraction, $(\beta,\kappa)=(0.1,50)$: the first coordinate needs only one step to meet the stated tolerance, yet Euler needs at least 25 steps to avoid amplifying the contracting second coordinate. Conversely, $(10,1)$ needs 25 steps for accuracy even though stability allows one.

\FloatBarrier
\subsection{Experiment 2: change stability without changing learned transport}
Experiment 1 uses a simple system with known dynamics. We now examine the same behavior using trained flow matching models. We train three independent checkpoints for each of two target distributions using standard independent-coupling flow matching. Eight-Gaussians contains eight isotropic Gaussian components arranged on a circle, while the bistable target is the finite-time law of an overdamped Langevin process in a double-well potential. We add one contracting coordinate to each trained model. Its contraction rate can be changed without affecting the original learned flow.
\begin{equation}
 \widetilde v_{\theta,\kappa}(s,(x,z))=(v_\theta(s,x),-\kappa z),
 \qquad
 \widetilde J=\operatorname{diag}(J_\theta,-\kappa).
\label{eq:augmentation}
\end{equation}

\begin{proposition}[Unchanged transport, independent stability floor]
\label{prop:augmentation}
The exact and Euler $x$-updates are unchanged by Eq.~\eqref{eq:augmentation}. On $z=0$, the original path and its material acceleration are unchanged, while the inserted mode requires $N\ge\lceil\kappa/2\rceil$ for non-expansion.
\end{proposition}

\begin{table}[t]
\centering\small
\caption{Controlled learned-flow augmentation at $\eps=0.05$, $\kappa=50$ (mean $\pm$ sample standard deviation over three seeds). The unaugmented checkpoints are accuracy-limited under the sampled local eigenmode test; the inserted mode then raises the joint requirement to 25 steps.}
\label{tab:controlled_key}
\begin{tabular}{@{}lrrrr@{}}\toprule
Target & Base accuracy & Base stability & Full error only & Joint requirement\\\midrule
Eight-Gaussians & $14.33\pm3.21$ & $4.00\pm0.00$ & $18.00\pm0.00$ & $25.00\pm0.00$\\
Bistable Langevin & $21.67\pm0.58$ & $1.67\pm0.58$ & $21.67\pm0.58$ & $25.00\pm0.00$\\\bottomrule
\end{tabular}
\end{table}

\begin{figure}[t]
\centering
\includegraphics[width=0.96\linewidth]{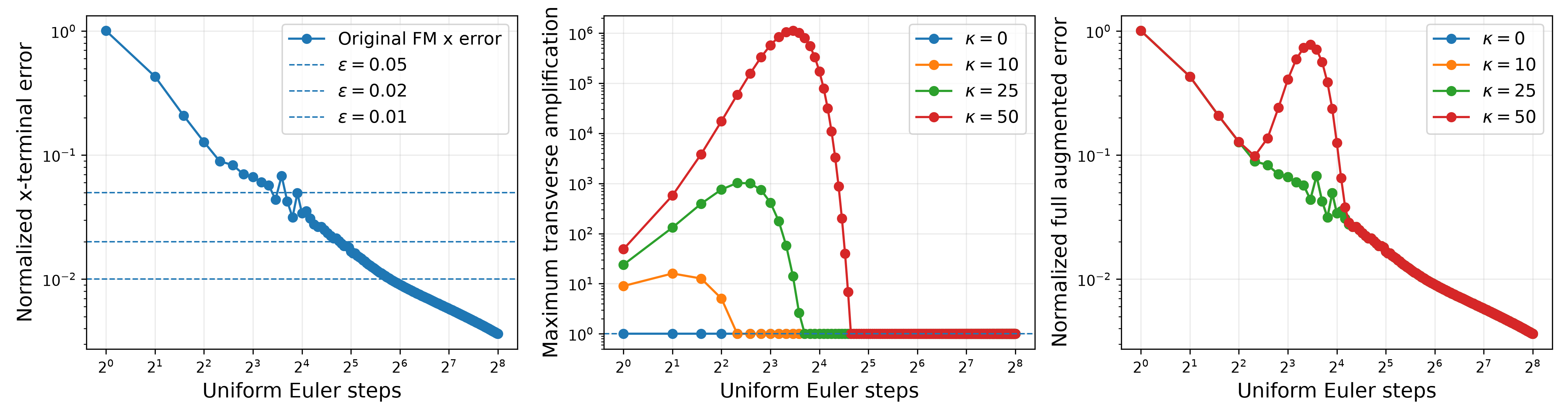}\\[-1mm]
\textbf{(a) Eight-Gaussians}\par\smallskip
\includegraphics[width=0.96\linewidth]{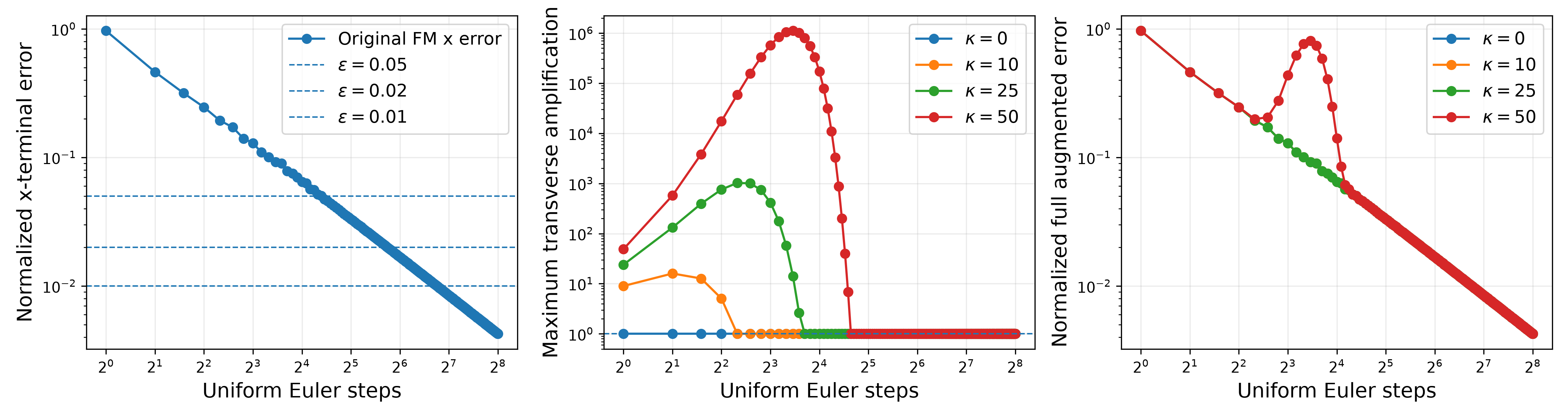}\\[-1mm]
\textbf{(b) Bistable Langevin}
\caption{\textbf{Experiment 2: controlled contraction added to trained flow matching models.} Representative seed-0 checkpoints are shown for (a) Eight-Gaussians and (b) bistable Langevin. Left: the original learned-coordinate Euler error is unchanged across $\kappa$. Middle: the inserted transverse direction is amplified below its exact Euler boundary $N=\lceil\kappa/2\rceil$. Right: the full augmented terminal error can become small before that transverse direction becomes non-expansive. Quantitative results over all three training seeds are reported in Table~\ref{tab:controlled_key}.}
\label{fig:exp2_seed0}
\end{figure}

Before augmentation, both learned flows are accuracy-limited at $\eps=0.05$. The accuracy budgets are $14.33\pm3.21$ steps for Eight-Gaussians and $21.67\pm0.58$ for Langevin, while their sampled stability budgets are only $4.00\pm0.00$ and $1.67\pm0.58$. Adding the contracting coordinate introduces exact stability floors of 5, 13, and 25 steps for $\kappa=10,25,50$, and all six checkpoints match these thresholds.

For $\kappa=50$, stability requires 25 Euler steps. However, the full endpoint error meets the 5\% tolerance at 18 steps for Eight-Gaussians and about 22 steps for Langevin (Table~\ref{tab:controlled_key}). Just below the stability boundary, the transverse amplification is $R_z^-=6.83$. This happens because the added coordinate starts very close to zero: its contribution to terminal error can remain small even while perturbations along that direction are strongly amplified.

Figure~\ref{fig:exp2_seed0} shows that changing the added contraction rate leaves the original learned flow unchanged while moving the Euler stability boundary. Thus, endpoint error alone can miss a separate stability requirement.

Whether accuracy or stability determines the step count also depends on the chosen error tolerance. At $\eps=0.02$ and $0.01$, the accuracy budgets increase to $27.67\pm2.52$ and $58.33\pm10.69$ for Eight-Gaussians, and $54.00\pm0.00$ and $108.33\pm1.15$ for Langevin. These values are already larger than the maximum stability floor of 25, so the same systems become accuracy-limited at stricter tolerances. We therefore describe a sampler as accuracy- or stability-limited only for a stated tolerance. The same behavior is reproduced by the remaining seeds in Figs.~\ref{fig:exp2_seed1}--\ref{fig:exp2_seed2}.

\FloatBarrier
Experiments 1 and 2 show that local stability can impose a step requirement separate from trajectory accuracy. They do not show that this happens in a pretrained robot policy. In Experiment 3, we therefore leave the $\pi_{0.5}$-LIBERO policy unchanged and test which requirement is active in its native action-flow field.

\section{Experiment 3: A Frozen $\pi_{0.5}$-LIBERO Policy} \label{sec:checkpoint}

\subsection{Local Stability Along the Sampling Trajectory} 

The Jacobian changes along the sampling trajectory and across policy inputs. To assess local Euler stability, we must compare its contracting modes with the step sizes used by the sampler. In OpenPI, Euler starts updates at
\begin{equation}
 t_k=1-k/N,\qquad k=0,\ldots,N-1,
\end{equation}
so it reaches $t=0$ but never starts a step there. \emph{A large Jacobian at the endpoint does not necessarily require more Euler steps because Euler never starts an update at that point}.

For a fixed reference path $x_c^{\rm ref}(t)$, define the largest Euler eigenmode amplification factor on that grid
\begin{equation}
 R_N^{\rm ref}(c)=\max_{0\le k<N}
 \max_{\substack{\mu\in\spec(J_v(t_k,x_c^{\rm ref}(t_k);c))\\\RePart(\mu)>0}}
 \left|1-\frac{\mu}{N}\right|.
\label{eq:grid_R}
\end{equation}
The condition $R_N^{\rm ref}\le1$ means that every contracting eigenmode included in the calculation is non-expansive under the corresponding frozen Euler update. It does not bound the product of time-varying update matrices.

To compare many possible budgets without recomputing every Euler grid, define the scaled stability measure
\begin{equation}
 G_c(t)=t\,\chi_{\leftarrow}(t,x_c^{\rm ref}(t);c),\qquad 0<t\le1.
\label{eq:G}
\end{equation}

\begin{proposition}[Sufficient reference-path condition]
\label{prop:grid}
If $G_c(t)\le1$ for every $t\in(0,1]$ when all Jacobian eigenvalues are included, then $R_N^{\rm ref}(c)\le1$ for every positive integer $N$.
\end{proposition}

\begin{proof}
Every Euler node can be written as $t_k=m/N$ for an integer $m\ge1$. If $t\chi_{\leftarrow}(t)\le1$, then $\chi_{\leftarrow}(t_k)\le1/t_k=N/m\le N$, which is exactly the local Euler non-expansion condition.
\end{proof}

Our calculations cover selected eigenvalues at a finite number of times. Therefore, finding no violation does not guarantee that the stability condition holds along the entire trajectory. Any detected violation should be checked directly on the corresponding Euler grid and with finite perturbations.

\subsection{Checkpoint and evaluated policy inputs}
We study the released \texttt{pi05\_libero} checkpoint without retraining, adding coordinates, or changing its learned field. The OpenPI JAX model integrates a $10\times32$ internal action tensor, giving a 320-dimensional ODE state, and transforms it to a $10\times7$ LIBERO action chunk \citep{openpi2026}. Thus each Jacobian is $320\times320$ and its eigenvectors are directions in the complete internal action-flow space.

We record the first three policy inputs from one default-policy rollout on each of the ten \texttt{libero\_10} tasks, giving 30 early-rollout policy inputs \citep{liu2023libero}. The tasks span diverse household manipulation, from placing multiple objects in baskets and arranging mugs on plates to operating a stove, placing a bowl in a drawer, loading a microwave, and placing a book in a caddy. Each saved policy input contains the visual observation used by the policy, the robot state, and the task instruction. In our saved data, each RGB image is a $224\times224\times3$ uint8 array and the robot state has eight float32 entries. The Gaussian $10\times32$ source-noise tensor is not included in the term \emph{policy input}; it is fixed separately for the numerical analysis, with source-noise seed 456 shared across all 30 inputs. The experiment therefore probes variation in policy conditioning while holding the analyzed noise draw fixed. Accuracy sweeps are complete on five inputs (the saved input labeled \texttt{context\_00} from tasks 00, 02, 04, 06, and 08), while all 30 have a reported maximum of the sampled stability measure.

\subsection{Numerical solver setup}
We measure Euler endpoint error against a refined RK4 solution using the same policy input and source noise. On the five paired cases, RK4-512/1024 endpoint disagreement is $3.37\times10^{-4}$ to $4.16\times10^{-4}$. We test every integer $N=1,\ldots,128$ and report the persistent threshold from Eq.~\eqref{eq:per_context_nacc}.

For stability, we evaluate an RK4-1024 reference trajectory at $t=1,31/32,\ldots,1/32$. At each sampled time, we use Jacobian–vector products and Arnoldi iteration to estimate six eigenvalues with the largest real part and six with the largest magnitude. We denote the largest computed value of $t\chi_{\leftarrow}$ by $\widehat G_{\max}$. Four full $320\times320$ JVP-assembled Jacobians check the Arnoldi eigenvalues at three states in task 00 and one state in task 06. The calculation does not include every eigenvalue at every state and does not cover the interval $0<t<1/32$.

\subsection{Results}

\begin{figure}[t]
\centering
\includegraphics[width=0.485\linewidth]{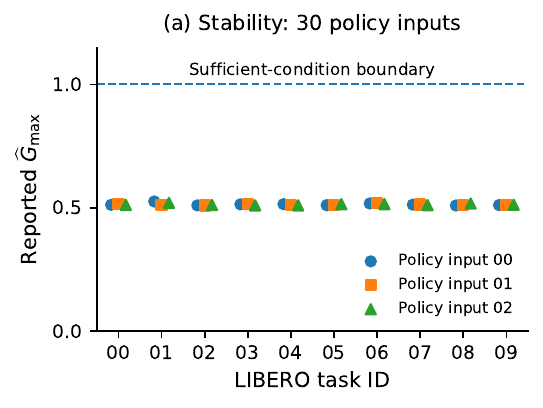}\hfill
\includegraphics[width=0.485\linewidth]{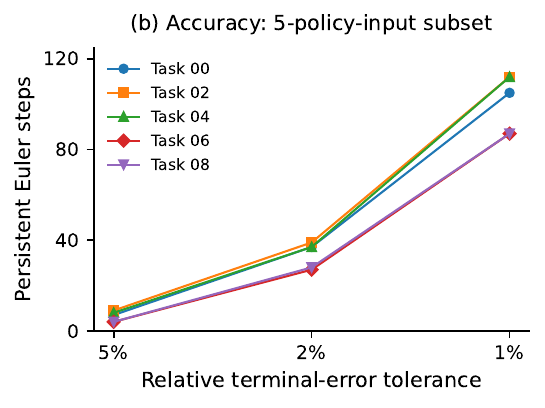}
\caption{\textbf{Experiment 3: Pretrained-policy accuracy and stability results.} Left: maximum sampled $G=t\chi_{\leftarrow}$ for 30 early-rollout policy inputs (three per task); $G=1$ marks the sufficient condition for local Euler non-expansion when all eigenvalues and times are covered. Right: persistent accuracy budgets for the five-input subset. The two panels have different coverage and should not be interpreted as a 30-input paired comparison.}
\label{fig:pi05}
\end{figure}

\textbf{No sampled contracting mode violates the Euler stability condition.}
Figure~\ref{fig:pi05} summarizes the sampled stability results across 30 policy inputs and the accuracy requirements for the five paired inputs. Across the 30 policy inputs, the reported $\widehat G_{\max}$ values lie in $[0.5088,0.5252]$, with mean $0.51285$, median $0.51181$, and sample standard deviation $0.00371$.
None crosses the sufficient-condition boundary $G=1$.
The largest value, $0.525207$, occurs for task 01 / input 00 at $t=1/32$.
The sampled stability measure varies little across the 30 inputs, despite differences in the observations, robot states, and task instructions. Thus, we find no sampled trajectory for which local Euler stability requires a smaller timestep under this criterion.
However, 25 of the 30 maxima occur at the smallest sampled time, $t=1/32$, showing that the strongest measured stability demand is usually near the action endpoint.

\textbf{Accuracy requires more steps as the tolerance tightens.}
On the five inputs with paired measurements, the required Euler steps increase from 4–9 at 5\% terminal error to 27–39 at 2\% and 87–112 at 1\% (Table~\ref{tab:pi05_accuracy}). Thus, all five inputs satisfy the 5\% criterion with fewer than the default 10 Euler steps, whereas substantially more steps are required once the numerical tolerance is tightened. Going from 5\% to 1\% increases the required budget by roughly an order of magnitude.
At the same time, the corresponding sampled stability values remain tightly clustered, with $\widehat G_{\max}\in[0.5088,0.5171]$ for these five inputs.
This separation suggests that the additional steps required at tighter tolerances are driven by trajectory accuracy rather than by a sampled Euler stability violation. Here, 1\% denotes relative error in the full internal action-flow tensor against RK4-1024, not a 1\% task-failure rate.

\begin{table}[t]
\centering\small
\caption{Paired accuracy and sampled-stability results on the five-input subset. $\widehat G_{\max}$ is the largest value over the Arnoldi eigenmodes and prescribed reference times.}
\label{tab:pi05_accuracy}
\begin{tabular}{@{}lrrrr@{}}\toprule
LIBERO task & $N_{\rm acc}(5\%)$ & $N_{\rm acc}(2\%)$ & $N_{\rm acc}(1\%)$ & $\widehat G_{\max}$\\\midrule
00 & 7 & 37 & 105 & 0.5115\\
02 & 9 & 39 & 112 & 0.5092\\
04 & 8 & 37 & 112 & 0.5142\\
06 & 4 & 27 & 87 & 0.5171\\
08 & 4 & 28 & 87 & 0.5088\\\bottomrule
\end{tabular}
\vspace{-4mm}
\end{table}

\textbf{Strong contraction is present, but the sampled modes do not violate the Euler stability condition.}
The values $\widehat G_{\max}<1$ do not mean that the policy Jacobian has weak eigenvalues.
For the fully assembled task-00 Jacobian, $\chi_{\leftarrow}$ increases from $0.51$ at $t=1$
to $1.01$ at $t=1/2$ and $16.41$ at $t=1/32$ (Table~\ref{tab:full_jacobian}).
At $t=1/32$, 319 of the 320 eigenvalues have positive real part, so contracting directions are widespread near the action endpoint.
Such modes can become unstable under explicit Euler when the timestep is too large.
Here, however, their contraction rates remain compatible with the tested Euler grid: the full Jacobian gives
$G=0.5127$ for task 00 and $0.5172$ for task 06, closely matching the Arnoldi values $0.5115$ and $0.5171$.

\textbf{Limits of the local stability test.}
Our stability test is local and based on frozen-time Jacobian eigenvalues. It does not capture possible transient growth caused by non-normal dynamics. If a violating mode is found, a direct perturbation test would provide stronger evidence by showing that a perturbation contracts under a refined solver but grows under coarse Euler sampling. Finally, numerical accuracy and local stability do not guarantee successful robot behavior; the learned policy itself may still produce a poor action.

\section{Conclusion}
\label{sec:conclusion}
Euler budgets in flow matching policies are often treated as a single accuracy knob, but the learned ODE contains two distinct numerical objects: the nominal trajectory and the local dynamics around it. Our controlled experiments show that trajectory accuracy and contracting-mode stability can impose different step requirements even when the original transport is unchanged. Jacobian eigenvalue analysis of a frozen $\pi_{0.5}$-LIBERO policy finds no sampled contracting mode that violates the Euler non-expansion condition across the 30 rollout inputs studied, while five paired accuracy sweeps become increasingly demanding as the tolerance tightens. The practical lesson is not that every flow policy is stiff, but that a fixed step budget should be interpreted through both the trajectory it follows and the local dynamics encountered along that trajectory.

\textbf{Limitations:} Our controlled experiments demonstrate that stability can impose an additional Euler step requirement, but we do not observe this limitation in the pretrained $\pi_{0.5}$-LIBERO policy. The policy analysis is limited to 30 early-rollout inputs with a single shared noise draw, and paired accuracy measurements are available for only five inputs. 

\textbf{Future work:} An important next step is to extend the analysis to more pretrained policies, noise draws, and complete robot rollouts. Denser Jacobian evaluations near the action endpoint and direct perturbation experiments could provide stronger evidence of whether pretrained policies encounter stability-limited regimes. It would also be useful to study how numerical accuracy and local stability affect the generated actions and downstream task success. 

\FloatBarrier
\bibliography{references}

@inproceedings{lipman2023flow,
  title={Flow Matching for Generative Modeling},
  author={Lipman, Yaron and Chen, Ricky T. Q. and Ben-Hamu, Heli and Nickel, Maximilian and Le, Matthew},
  booktitle={International Conference on Learning Representations},
  year={2023}
}

@article{dai2025actionflow,
  title={ActionFlow: A Pipelined Action Acceleration for Vision Language Models on Edge},
  author={Dai, Yuntao and Gu, Hang and Wang, Teng and Cheng, Qianyu and Zheng, Yifei and Qiu, Zhiyong and Gong, Lei and Lou, Wenqi and Zhou, Xuehai},
  journal={arXiv preprint arXiv:2512.20276},
  year={2025}
}

@article{black2025pi05,
  title={$\backslash$pi\_ $\{$0.5$\}$: a Vision-Language-Action Model with Open-World Generalization},
  author={{Physical Intelligence} and Black, Kevin and Brown, Noah and Darpinian, James and Dhabalia, Karan and Driess, Danny and Esmail, Adnan and Equi, Michael and Finn, Chelsea and Fusai, Niccolo and others},
  journal={arXiv preprint arXiv:2504.16054},
  year={2025}
}

@inproceedings{chen2018neural,
  title={Neural Ordinary Differential Equations},
  author={Chen, Ricky T. Q. and Rubanova, Yulia and Bettencourt, Jesse and Duvenaud, David},
  booktitle={Advances in Neural Information Processing Systems},
  volume={31}, year={2018}
}

@book{hairer1996solving,
  title={Solving Ordinary Differential Equations II: Stiff and Differential-Algebraic Problems},
  author={Hairer, Ernst and Wanner, Gerhard},
  publisher={Springer}, year={1996}
}

@book{trefethen2005spectra,
  title={Spectra and Pseudospectra: The Behavior of Nonnormal Matrices and Operators},
  author={Trefethen, Lloyd N. and Embree, Mark},
  publisher={Princeton University Press}, year={2005}
}

@article{gupta2026sharpen,
  title={Sharpen Your Flow: Sharpness-Aware Sampling for Flow Matching},
  author={Gupta, Aditi and Lim, Soon Hoe and Yu, Annan and Erichson, N. Benjamin},
  journal={arXiv preprint arXiv:2605.11547}, year={2026},
  url={https://arxiv.org/abs/2605.11547}
}

@inproceedings{tan2026stork,
  title={{STORK}: Faster Diffusion and Flow Matching Sampling by Resolving Both Stiffness and Structure-Dependence},
  author={Tan, Zheng and Wang, Weizhen and Bertozzi, Andrea L. and Ryu, Ernest K.},
  booktitle={International Conference on Learning Representations}, year={2026}
}

@article{tao2026strain,
  title={On the Role of Strain and Vorticity in Numerical Integration Error for Flow Matching},
  author={Tao, Chenxi and Choi, Seung-Kyum},
  journal={arXiv preprint arXiv:2605.06680}, year={2026},
  url={https://arxiv.org/abs/2605.06680}
}

@article{xiao2026euler,
  title={From {Euler} to {Dormand--Prince}: {ODE} Solvers for Flow Matching Generative Models},
  author={Xiao, Hao}, journal={arXiv preprint arXiv:2605.00836}, year={2026},
  url={https://arxiv.org/abs/2605.00836}
}

@article{tong2024improving,
  title={Improving and Generalizing Flow-Based Generative Models with Minibatch Optimal Transport},
  author={Tong, Alexander and Fatras, Kilian and Malkin, Nikolay and Huguet, Guillaume and Zhang, Yanlei and Rector-Brooks, Jarrid and Wolf, Guy and Bengio, Yoshua},
  journal={Transactions on Machine Learning Research}, year={2024}
}

@inproceedings{kornilov2024optimal,
  title={Optimal Flow Matching: Learning Straight Trajectories in Just One Step},
  author={Kornilov, Nikita and Mokrov, Petr and Gasnikov, Alexander and Korotin, Alexander},
  booktitle={Advances in Neural Information Processing Systems}, year={2024}
}

@article{black2024pi0,
  title={{$\pi_0$}: A Vision-Language-Action Flow Model for General Robot Control},
  author={Black, Kevin and Brown, Noah and Driess, Danny and Esmail, Adnan and Equi, Michael and Finn, Chelsea and others},
  journal={arXiv preprint arXiv:2410.24164}, year={2024},
  url={https://arxiv.org/abs/2410.24164}
}

@inproceedings{chi2023diffusion,
  title={Diffusion Policy: Visuomotor Policy Learning via Action Diffusion},
  author={Chi, Cheng and Feng, Siyuan and Du, Yilun and Xu, Zhenjia and Cousineau, Eric and Burchfiel, Benjamin C. M. and Song, Shuran},
  booktitle={Robotics: Science and Systems}, year={2023},
  doi={10.15607/RSS.2023.XIX.026}
}

@inproceedings{liu2023libero,
  title={{LIBERO}: Benchmarking Knowledge Transfer for Lifelong Robot Learning},
  author={Liu, Bo and Zhu, Yifeng and Gao, Chongkai and Feng, Yihao and Liu, Qiang and Zhu, Yuke and Stone, Peter},
  booktitle={Advances in Neural Information Processing Systems}, year={2023},
  url={https://arxiv.org/abs/2306.03310}
}

@article{fang2026probeflow,
  title={{ProbeFlow}: Training-Free Adaptive Flow Matching for Vision-Language-Action Models},
  author={Fang, Zhou and Wang, Jiaqi and Zhou, Yi and Shi, Qiongfeng},
  journal={arXiv preprint arXiv:2603.17850}, year={2026},
  url={https://arxiv.org/abs/2603.17850}
}

@article{tang2026coda,
  title={Fewer Steps, Better Actions: Rethinking Flow-Matching Inference for {VLA} Policies},
  author={Tang, Zhipeng and Chen, Xinda and Rao, Weining and Li, Xiao and Tan, Wenting and Wang, Yuning and Shi, Xiao and Zhao, Xiaofang},
  journal={arXiv preprint arXiv:2609.21216}, year={2026},
  url={https://arxiv.org/abs/2609.21216}
}

@misc{openpi2026,
  title={{OpenPI}: Open-Source Models for Robotics},
  author={{Physical Intelligence}}, year={2026},
  howpublished={Software repository},
  note={Implementation inspected at commit 215abfb217dbac7d5f1273282331b9b1866c0479},
  url={https://github.com/Physical-Intelligence/openpi}
}

@inproceedings{hu2024adaflow,
  title={{AdaFlow}: Imitation Learning with Variance-Adaptive Flow-Based Policies},
  author={Hu, Xixi and Liu, Bo and Liu, Xingchao and Liu, Qiang},
  booktitle={Advances in Neural Information Processing Systems},
  volume={37}, year={2024},
  url={https://proceedings.neurips.cc/paper_files/paper/2024/hash/fa9cfdb49f7de3dee54007f84c0745b1-Abstract-Conference.html}
}

@inproceedings{zhang2025flowpolicy,
  title={{FlowPolicy}: Enabling Fast and Robust {3D} Flow-Based Policy via Consistency Flow Matching for Robot Manipulation},
  author={Zhang, Qinglun and Liu, Zhen and Fan, Haoqiang and Liu, Guanghui and Zeng, Bing and Liu, Shuaicheng},
  booktitle={Proceedings of the AAAI Conference on Artificial Intelligence},
  volume={39}, number={14}, pages={14754--14762}, year={2025},
  doi={10.1609/aaai.v39i14.33617},
  url={https://ojs.aaai.org/index.php/AAAI/article/view/33617}
}

\clearpage
\appendix

\section{Proofs and Details}
\label{app:proofs}

\subsection{Proof of Theorem~\ref{thm:exact}}
Let $h=1/N$ and $s_k=kh$. The exact increment of the first coordinate is
\[
 x_1(s_k+h)-x_1(s_k)=h(1+\beta s_k)+\frac{\beta h^2}{2}.
\]
Euler omits the last term. Summing over $N$ steps gives
\[
 x_1(1)-\widehat x_{1,N}=N\frac{\beta h^2}{2}=\frac{\beta}{2N}.
\]
Therefore the terminal-error condition holds exactly when $N\ge\lceil\beta/(2\eps)\rceil$. The second coordinate follows
\[
 \widehat x_{2,k+1}=(1-\kappa/N)\widehat x_{2,k},
\]
which is non-expansive exactly when $|1-\kappa/N|\le1$, or $N\ge\lceil\kappa/2\rceil$. The joint minimum is the maximum of the two requirements. Equality at $\kappa/N=2$ gives neutral sign oscillation; the theorem requires non-expansion, not accurate resolution of the decay.

\subsection{Proof of Proposition~\ref{prop:augmentation}}
The augmented Jacobian is block diagonal, so
\begin{equation}
 \spec(\widetilde J)=\spec(J_\theta)\cup\{-\kappa\},\qquad
 \chi(\widetilde J)=\max\{\chi(J_\theta),\kappa/2\}.
\end{equation}
Because the added coordinate does not feed back into $x$, both exact and Euler $x$-trajectories are unchanged. The scalar coordinate satisfies
\[
 z(1)=e^{-\kappa}z_0,\qquad
 \widehat z_N=(1-\kappa/N)^N z_0.
\]
On the invariant plane $z=0$, the added material acceleration vanishes, so the original path and its material acceleration do not reveal the independently adjustable contraction.

\subsection{Why the stability measure uses sampled times}
An endpoint-inclusive Jacobian maximum is not equivalent to an Euler-grid test. Proposition~\ref{prop:grid} uses the fact that every actual update starts at some $t=m/N>0$. If sampled eigenvalue calculations give $G(t)<1$, then no violation was found among those eigenmodes and times; this does not establish the proposition's condition for every time and every Jacobian eigenvalue.

\newpage
\section{Controlled-Experiment Setup}
\label{app:controlled}

\subsection{Experiment 1}
We use 1000 initial conditions with $x_1(0)\sim\mathcal N(0,1)$ and $x_2(0)\sim\mathcal N(0,10^{-12})$, tolerance $\eps=0.2$, and every integer $N=1,\ldots,128$. The four parameter pairs are $\{0.1,10\}\times\{1,50\}$. For nonzero initial perturbations, the transverse amplification is
\begin{equation}
 R_z(N)=\max_{0\le k\le N}\frac{|\widehat x_{2,k}|}{|x_2(0)|}
       =\max\{1,|1-\kappa/N|^N\},
\end{equation}
so $R_z(N)>1$ exactly when $N<\kappa/2$.

\subsection{Experiment 2: targets and training}
The Eight-Gaussians target is an eight-component isotropic Gaussian mixture with means equally spaced on a circle of radius 4 and component standard deviation 0.25 \citep{tong2024improving,kornilov2024optimal}. The bistable Langevin target is the finite-time law of
\[
 dY_u=-\nabla U(Y_u)\,du+\sqrt{2\tau}\,dW_u,\qquad
 U(y)=\tfrac14(y_1^2-1)^2+\tfrac12y_2^2,
\]
with $\tau=0.25$, $Y_0\sim\mathcal N(0,I_2)$, Euler--Maruyama step $10^{-3}$, and terminal time $u=4$. The dataset contains $100{,}000$ training and $10{,}000$ held-out samples.

Source and target endpoints are sampled independently, with $X_0\sim\mathcal N(0,I_2)$ and linear flow matching interpolation. Each model uses four hidden layers of width 128, SiLU activations, and a 32-dimensional sinusoidal time embedding. We train three seeds per target for 50,000 Adam steps with batch size 512 and learning rate $10^{-3}$.

For 500 matched held-out source samples, the learned ODE is solved in float64 with batched RK4 at 6400 steps; a fixed subset recomputed at 12800 steps has maximum normalized disagreement $5.3\times10^{-14}$. We evaluate $\kappa\in\{0,10,25,50\}$, every integer $N=1,\ldots,256$, and $\eps\in\{0.05,0.02,0.01\}$.

\begin{table}[H]
\centering\scriptsize
\caption{Complete Experiment 2 results at $\eps=0.05$. $N_{\rm base}$ is the persistent accuracy threshold of the unaugmented learned field, $\widehat N_{\rm stab}^{\rm base}$ is its sampled local-eigenmode stability budget, and $N_{\rm insert}=\lceil\kappa/2\rceil$ is the exact inserted stability floor.}
\label{tab:full_controlled}
\resizebox{\linewidth}{!}{%
\begin{tabular}{@{}lrrrrrrr@{}}\toprule
Target & $\kappa$ & $N_{\rm base}$ & $\widehat N_{\rm stab}^{\rm base}$ & $N_{\rm insert}$ & $N_{\rm pred}=N_{\rm joint}$ & $N_{\rm full}$ & $R_z^-$\\\midrule
Eight-Gaussians & 0  & $14.33\pm3.21$ & $4.00\pm0.00$ & 0  & $14.33\pm3.21$ & $14.33\pm3.21$ & --\\
 & 10 & $14.33\pm3.21$ & $4.00\pm0.00$ & 5  & $14.33\pm3.21$ & $14.33\pm3.21$ & 5.06\\
 & 25 & $14.33\pm3.21$ & $4.00\pm0.00$ & 13 & $14.67\pm2.89$ & $14.33\pm3.21$ & 2.61\\
 & 50 & $14.33\pm3.21$ & $4.00\pm0.00$ & 25 & $25.00\pm0.00$ & $18.00\pm0.00$ & 6.83\\\midrule
Bistable Langevin & 0  & $21.67\pm0.58$ & $1.67\pm0.58$ & 0  & $21.67\pm0.58$ & $21.67\pm0.58$ & --\\
 & 10 & $21.67\pm0.58$ & $1.67\pm0.58$ & 5  & $21.67\pm0.58$ & $21.67\pm0.58$ & 5.06\\
 & 25 & $21.67\pm0.58$ & $1.67\pm0.58$ & 13 & $21.67\pm0.58$ & $21.67\pm0.58$ & 2.61\\
 & 50 & $21.67\pm0.58$ & $1.67\pm0.58$ & 25 & $25.00\pm0.00$ & $21.67\pm0.58$ & 6.83\\\bottomrule
\end{tabular}%
}
\end{table}
Here $R_z^-$ is the transverse amplification immediately below the inserted stability boundary. The observed inserted-mode boundaries are 5, 13, and 25 for $\kappa=10,25,50$ on all six checkpoints. At $\kappa=50$, the joint threshold rises to 25 even though the full terminal-error threshold remains smaller; immediately below the boundary, $R_z^-=6.83$. Across all 72 seed--tolerance--$\kappa$ combinations, the observed joint threshold matches $\max\{N_{\rm base},N_{\rm insert}\}$, which is a consistency check of the controlled decomposition rather than an out-of-sample predictor.

At stricter tolerances, the unaugmented accuracy budgets already exceed the largest inserted floor: $27.67\pm2.52$ and $58.33\pm10.69$ for Eight-Gaussians, and $54.00\pm0.00$ and $108.33\pm1.15$ for Langevin at $\eps=0.02$ and $0.01$, respectively. Thus the $\kappa=50$ augmented systems that are stability-limited at $\eps=0.05$ become accuracy-limited once the requested trajectory tolerance is tightened.

\newpage
\section{Additional Controlled Results}
\label{app:controlled_results}

The main text shows the exact mechanism plot and one representative trained-flow checkpoint. This appendix provides the target distributions and the remaining independently trained seeds used in the reported means and standard deviations.

\begin{figure}[h]
\centering
\begin{minipage}{.48\linewidth}\centering
\includegraphics[width=\linewidth]{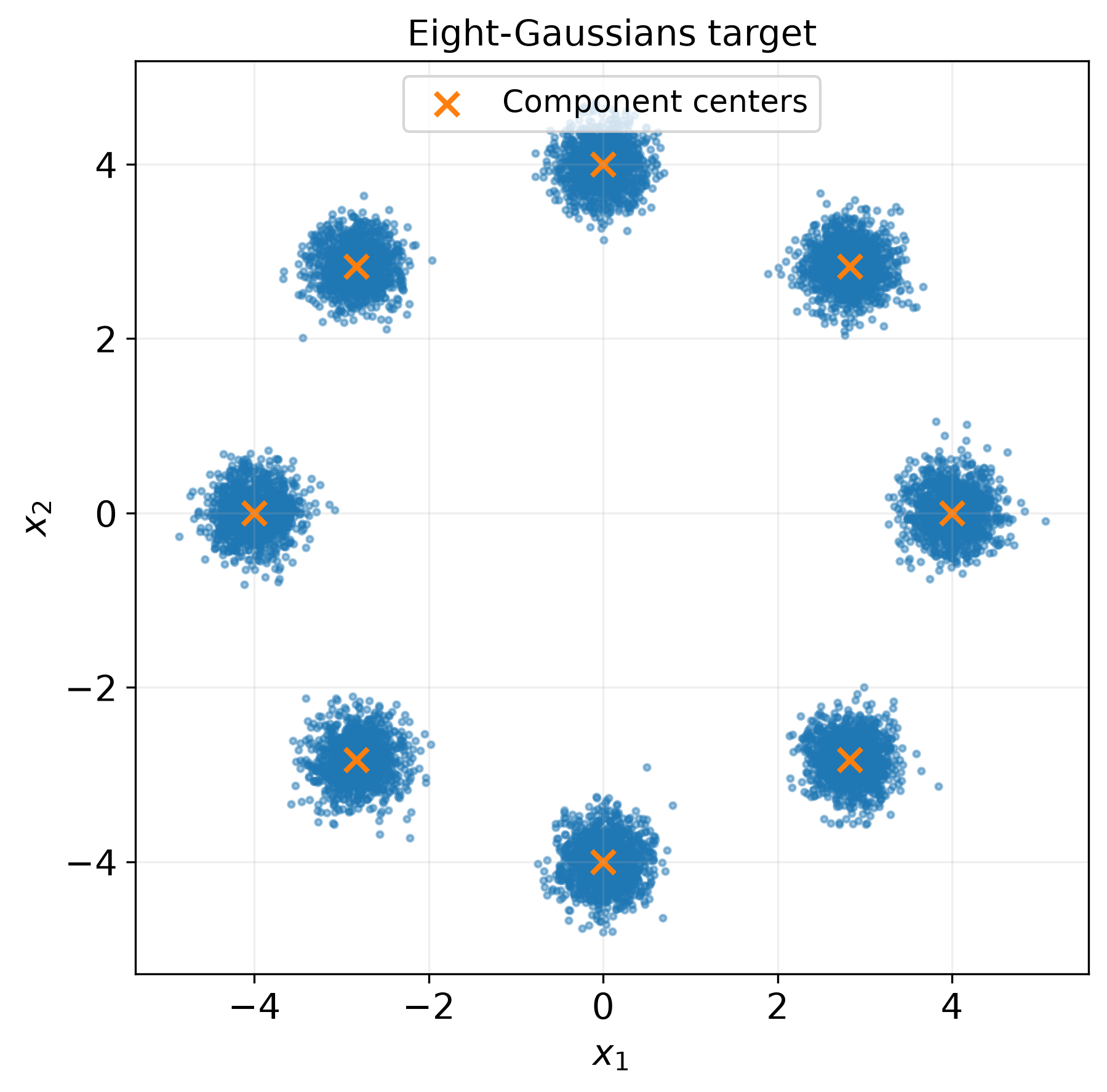}\\(a) Eight-Gaussians
\end{minipage}\hfill
\begin{minipage}{.48\linewidth}\centering
\includegraphics[width=\linewidth]{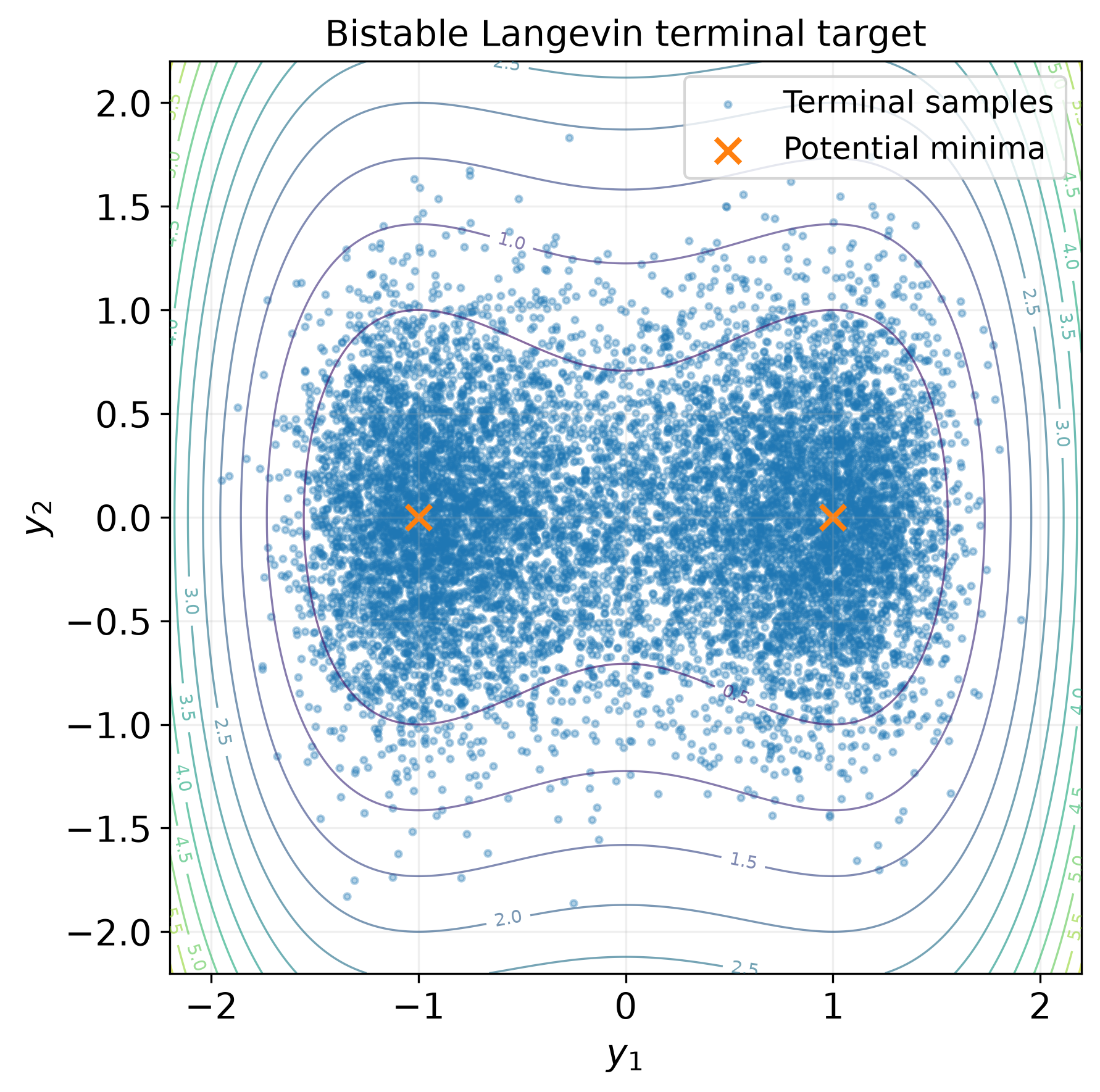}\\(b) Bistable Langevin
\end{minipage}
\caption{Two-dimensional targets used in Experiment 2. Eight-Gaussians contains eight isotropic components arranged on a circle of radius 4 with component standard deviation 0.25. Bistable Langevin is the finite-time distribution generated by the double-well dynamics specified in Appendix~\ref{app:controlled}. Both targets are standardized before flow matching training.}
\label{fig:2Ddatasets}
\end{figure}

\begin{figure}[h]
\centering
\includegraphics[width=\linewidth]{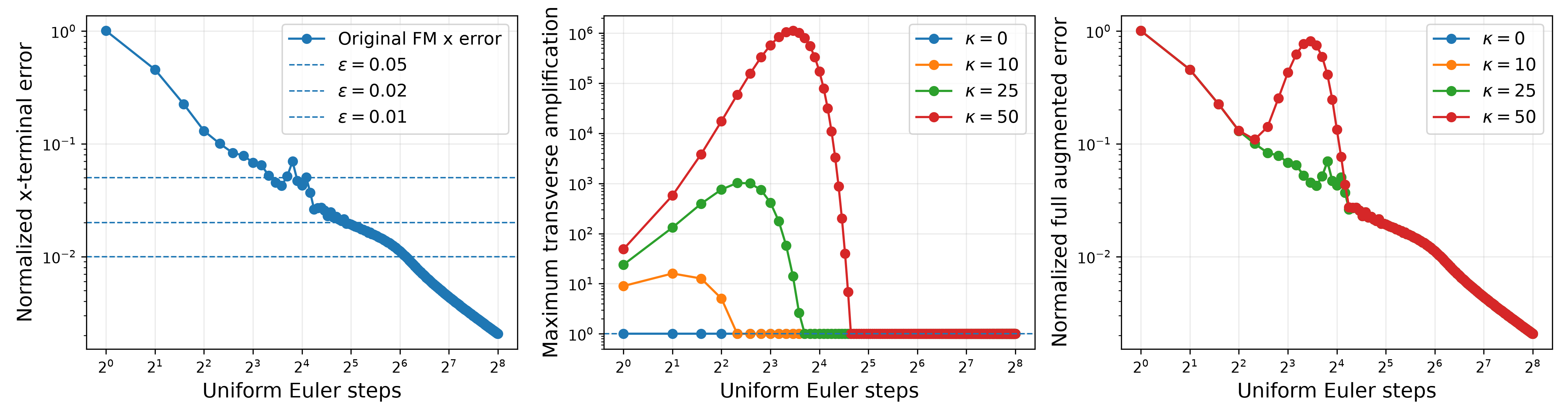}\\
\textbf{(a) Eight-Gaussians}\par\medskip
\includegraphics[width=\linewidth]{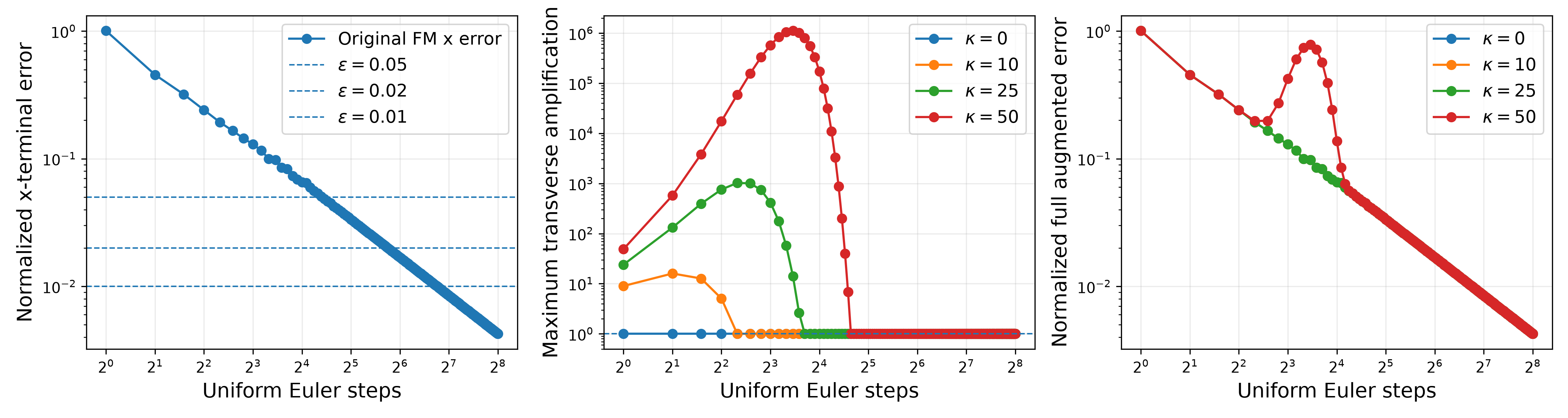}\\
\textbf{(b) Bistable Langevin}
\caption{Experiment 2, seed 1. As in the main-text representative, changing $\kappa$ leaves the original learned-coordinate error unchanged (left), moves only the transverse amplification boundary (middle), and can leave the full augmented terminal error below tolerance even before transverse non-expansion is reached (right).}
\label{fig:exp2_seed1}
\end{figure}

\begin{figure}[h]
\centering
\includegraphics[width=\linewidth]{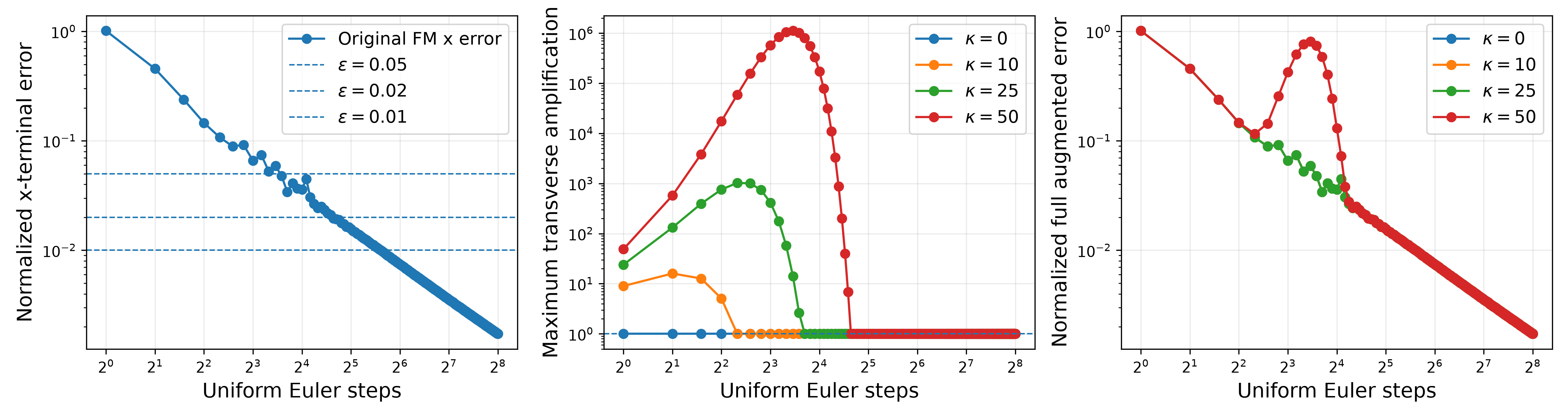}\\
\textbf{(a) Eight-Gaussians}\par\medskip
\includegraphics[width=\linewidth]{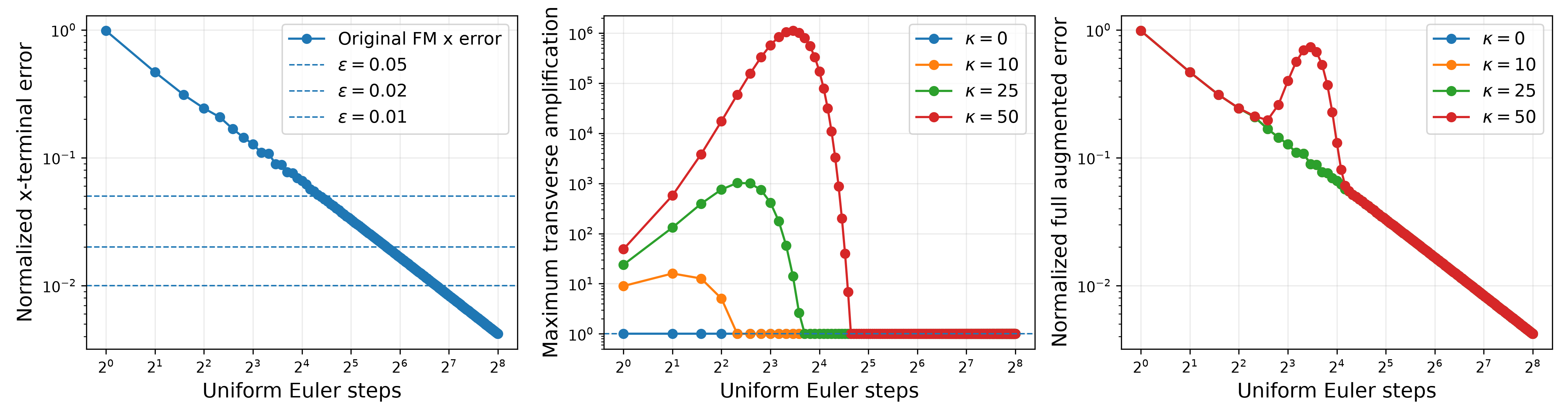}\\
\textbf{(b) Bistable Langevin}
\caption{Experiment 2, seed 2. The same separation is reproduced in the third independently trained checkpoint for each target. These two appendix figures, together with Fig.~\ref{fig:exp2_seed0}, underlie the three-seed summary in Tables~\ref{tab:controlled_key} and~\ref{tab:full_controlled}.}
\label{fig:exp2_seed2}
\end{figure}

The agreement across seeds is important because the inserted scalar mode is analytically known while the nominal two-dimensional velocity fields are independently learned. The stability boundary therefore tracks the controlled parameter $\kappa$, whereas differences across checkpoints appear in the baseline trajectory-accuracy requirement.

\clearpage
\section{Pretrained-Policy Numerical Details and Checks}
\label{app:checkpoint}

\subsection{Execution and implementation checks}
The completed runs use one NVIDIA H100 with 80 GB memory on Linux, JAX 0.5.3, and NumPy 1.26.4. LIBERO runs in a separate Python 3.8 environment; the model environment uses Python 3.11. MuJoCo observations are rendered with OSMesa on CPU. No real robot was used.

The checkpoint is \texttt{gs://openpi-assets/checkpoints/pi05\_libero}; normalization statistics come from its \texttt{assets} directory. The upstream code base is OpenPI commit \texttt{215abfb217dbac7d5f1273282331b9b1866c0479}, with only server-side argument forwarding added to expose the existing Euler step count. With a fixed observation and fixed $10\times32$ Gaussian noise, default sampling and explicit $N=10$ produce bitwise-identical $10\times7$ returned actions.

\subsection{Reference refinement}
Task 00 received the deeper RK4 refinement sequence below.
\begin{table}[H]
\centering\small
\caption{Task-00 RK4 endpoint refinement. Values are normalized by the finer endpoint norm.}
\begin{tabular}{@{}rrr@{}}\toprule
Coarse steps & Fine steps & Relative endpoint disagreement\\\midrule
16 & 32 & $3.3481\times10^{-2}$\\
32 & 64 & $1.7390\times10^{-2}$\\
64 & 128 & $5.4245\times10^{-3}$\\
128 & 256 & $3.2001\times10^{-3}$\\
256 & 512 & $9.6617\times10^{-4}$\\
512 & 1024 & $3.7975\times10^{-4}$\\\bottomrule
\end{tabular}
\end{table}
Across the five paired policy inputs, RK4-512/1024 disagreement ranges from $3.3697\times10^{-4}$ to $4.1593\times10^{-4}$. These are empirical refinement checks, not rigorous error bounds on the unknown exact solution.

\subsection{Euler-grid eigenvalue checks}
An initial task-00 calculation that included $t=0$ found a local stability scale near 496 at the endpoint. Because no Euler step starts at $t=0$, this value is not a sampler budget and motivated the grid-aware analysis in Section~\ref{sec:checkpoint}. For task 00, the maxima computed over reference-state Euler grids are:
\begin{table}[H]
\centering\small
\caption{Task-00 candidate maxima $\widehat R_N$ on the reference-state Euler grids.}
\begin{tabular}{@{}rrrrrrrrrrr@{}}\toprule
$N$ & 1 & 2 & 3 & 4 & 5 & 6 & 7 & 8 & 9 & 10\\\midrule
$\widehat R_N$ & .0098 & .4964 & .6641 & .7480 & .7984 & .8322 & .8559 & .8739 & .8879 & .8992\\\bottomrule
\end{tabular}
\end{table}

The 30-input stability statistics are descriptive for a common-noise, early-rollout collection. Adjacent policy inputs within a task are correlated, so the mean and sample standard deviation are not a population confidence interval.

{\small
\begin{longtable}{@{}llrrrrrr@{}}
\caption{Maximum sampled $G=t\chi_{\leftarrow}$ for all 30 policy inputs. $r_7$ is the norm fraction in the first seven internal coordinates of each action token.}\label{tab:all_stability}\\
\toprule
Task & Input & $\widehat G_{\max}$ & $t_{\max}$ & $\chi$ & Residual & $r_\perp$ & $r_7$\\
\midrule
\endfirsthead
\multicolumn{8}{l}{\textit{Table \ref{tab:all_stability} continued}}\\
\toprule
Task & Input & $\widehat G_{\max}$ & $t_{\max}$ & $\chi$ & Residual & $r_\perp$ & $r_7$\\
\midrule
\endhead
00 & 00 & 0.511522 & 0.03125 & 16.36869 & 0.00520 & 0.99978 & 0.88609\\
00 & 01 & 0.513789 & 0.03125 & 16.44125 & 0.00446 & 0.99997 & 0.90848\\
00 & 02 & 0.511219 & 0.03125 & 16.35901 & 0.00549 & 0.99968 & 0.28063\\
01 & 00 & 0.525207 & 0.03125 & 16.80663 & 0.00463 & 0.99972 & 0.96933\\
01 & 01 & 0.511849 & 0.03125 & 16.37918 & 0.00612 & 0.99991 & 0.43778\\
01 & 02 & 0.519417 & 0.03125 & 16.62134 & 0.00458 & 0.99996 & 0.97099\\
02 & 00 & 0.509235 & 0.09375 & 5.43184 & 0.00525 & 0.99992 & 0.31005\\
02 & 01 & 0.509391 & 0.03125 & 16.30052 & 0.00568 & 0.99947 & 0.14934\\
02 & 02 & 0.510954 & 0.03125 & 16.35052 & 0.00581 & 0.99996 & 0.22800\\
03 & 00 & 0.513597 & 0.03125 & 16.43509 & 0.00527 & 0.99984 & 0.66583\\
03 & 01 & 0.514275 & 0.09375 & 5.48560 & 0.00520 & 0.99939 & 0.86231\\
03 & 02 & 0.509332 & 0.09375 & 5.43287 & 0.00509 & 0.99973 & 0.58325\\
04 & 00 & 0.514207 & 0.03125 & 16.45462 & 0.00506 & 0.99990 & 0.92801\\
04 & 01 & 0.509913 & 0.03125 & 16.31721 & 0.00523 & 1.00000 & 0.52608\\
04 & 02 & 0.509332 & 0.03125 & 16.29861 & 0.00537 & 0.99989 & 0.13519\\
05 & 00 & 0.510091 & 0.71875 & 0.70969 & 0.00455 & 0.94757 & 0.27311\\
05 & 01 & 0.511767 & 0.03125 & 16.37655 & 0.00631 & 0.99978 & 0.21229\\
05 & 02 & 0.514214 & 0.03125 & 16.45485 & 0.00484 & 0.99999 & 0.95263\\
06 & 00 & 0.517069 & 0.03125 & 16.54622 & 0.00555 & 1.00000 & 0.91189\\
06 & 01 & 0.519674 & 0.03125 & 16.62957 & 0.00471 & 0.99963 & 0.97700\\
06 & 02 & 0.514482 & 0.03125 & 16.46341 & 0.00551 & 0.99999 & 0.92091\\
07 & 00 & 0.512238 & 0.03125 & 16.39163 & 0.00522 & 0.99988 & 0.90050\\
07 & 01 & 0.512495 & 0.03125 & 16.39985 & 0.00486 & 0.99991 & 0.68072\\
07 & 02 & 0.510317 & 0.03125 & 16.33016 & 0.00528 & 1.00000 & 0.91113\\
08 & 00 & 0.508802 & 0.75000 & 0.67840 & 0.00508 & 0.93454 & 0.37949\\
08 & 01 & 0.510430 & 0.03125 & 16.33376 & 0.00556 & 0.99952 & 0.46242\\
08 & 02 & 0.516805 & 0.03125 & 16.53777 & 0.00566 & 0.99967 & 0.27136\\
09 & 00 & 0.510554 & 0.03125 & 16.33772 & 0.00715 & 0.99998 & 0.09263\\
09 & 01 & 0.511193 & 0.03125 & 16.35816 & 0.00552 & 0.99988 & 0.37491\\
09 & 02 & 0.512045 & 0.03125 & 16.38543 & 0.00504 & 0.99981 & 0.97094\\
\bottomrule
\end{longtable}
}

\begin{figure}[H]
\centering
\includegraphics[width=0.82\linewidth]{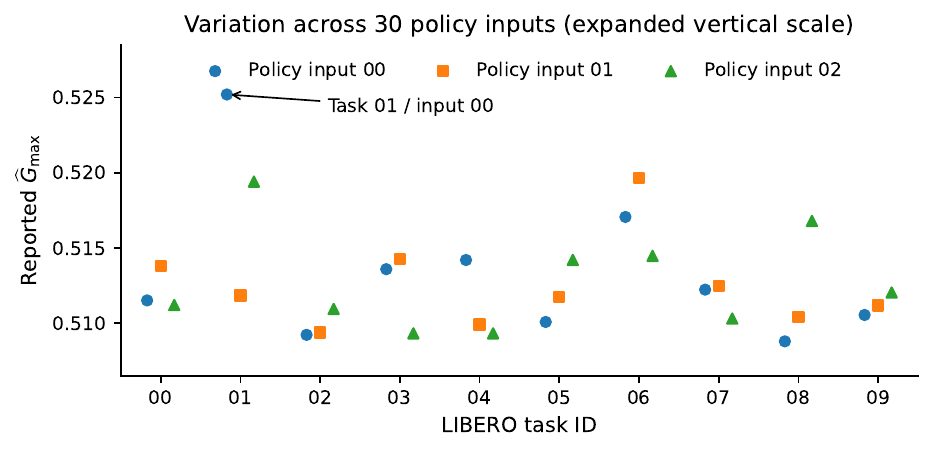}
\caption{Expanded view of the maximum sampled stability measure for the 30 policy inputs. The sufficient boundary $G=1$ lies outside this restricted vertical range and is shown in Figure~\ref{fig:pi05}.}
\end{figure}

\subsection{Full-Jacobian eigenvalue checks and geometry}
Four complete matrices were assembled from basis-direction JVPs and eigendecomposed after conversion to float64. The comparison with Arnoldi is:

\begin{table}[H]
\centering\small
\caption{Arnoldi versus full-Jacobian local stability scales.}
\label{tab:full_jacobian}
\begin{tabular}{@{}llrrr@{}}\toprule
Input & $t$ & Arnoldi $\chi$ & Full Jacobian $\chi$ & Relative difference\\\midrule
Task 00 / 00 & $1$ & 0.50535361 & 0.50603892 & 0.1354\%\\
Task 00 / 00 & $1/2$ & 1.01024206 & 1.01169466 & 0.1436\%\\
Task 00 / 00 & $1/32$ & 16.36869355 & 16.40657338 & 0.2309\%\\
Task 06 / 00 & $1/32$ & 16.54621714 & 16.55066042 & 0.0268\%\\\bottomrule
\end{tabular}
\end{table}

At task 06, the complete spectrum gives $G=0.51720814<1$, compared with the Arnoldi value $0.51706929$. The task-00 matrix at $t=1/32$ has 319 positive-real-part eigenvalues. These are local frozen-time eigenmodes, not proof that every finite-time perturbation contracts.

Across the 30 maximizing modes, the fraction of mode norm orthogonal to the local tangent ranges from approximately 0.9345 to 1, while the fraction in the first seven internal coordinates of each action token ranges from 0.0926 to 0.9770. Near-orthogonality alone is not exceptional in 320 dimensions; the contraction rate relative to the timestep is the important quantity.

\section{Notations}
\label{app:notation}
\begin{table}[H]
\centering\small
\caption{Main quantities used in the paper.}
\begin{tabularx}{\linewidth}{@{}lX@{}}\toprule
Symbol & Meaning\\\midrule
$c$ & Fixed policy input: visual observation, robot state (proprioception), and language instruction.\\
$\xi$ & Fixed Gaussian $10\times32$ source-noise tensor that initializes the action-flow ODE; treated separately from the policy input $c$.\\
$t$ & Native OpenPI sampling time: $1$ (noise) to $0$ (action).\\
$J_v$ & State Jacobian of the native-time velocity at fixed $c,t$.\\
$E_N(c,\xi)$ & Terminal error for one conditioning input and source-noise draw relative to the refined endpoint.\\
$N_{\rm acc}$ & Persistent trajectory-accuracy threshold within a finite tested budget set.\\
$\chi_{\leftarrow}$ & Local reverse-time Euler stability scale for contracting modes.\\
$R_N^{\rm ref}$ & Frozen-mode maximum on a specified reference-state Euler grid.\\
$G(t)=t\chi_{\leftarrow}$ & Scaled stability measure; $G\le1$ gives the sufficient reference-path condition when all times and eigenvalues are included.\\
$r_\perp$ & Mode norm fraction orthogonal to the local internal-flow tangent.\\
$r_7$ & Mode norm fraction in the first seven internal coordinates per token.\\\bottomrule
\end{tabularx}
\end{table}

\end{document}